\documentclass[10pt,twocolumn,letterpaper]{article}

\usepackage{cvpr}              

\usepackage{graphicx}
\usepackage{amsmath}
\usepackage{amssymb}
\usepackage{booktabs}
\usepackage{cite}
\usepackage{amsmath,amssymb,amsfonts}
\usepackage{algorithmic}
\usepackage{graphicx}
\usepackage{textcomp}
\usepackage{cite}
\usepackage{multirow}
\usepackage{multicol}
\usepackage{cite}
\usepackage{amsmath}
\usepackage{caption}
\usepackage{algorithm}
\usepackage{algorithmic}
\usepackage{kotex}
\usepackage{tabularx}
\usepackage{multirow}
\usepackage{adjustbox}
\usepackage{amssymb}
\usepackage{commath}
\usepackage{booktabs}

\newcommand{\red}[1]{\textcolor{red}{\textbf{#1}}}
\newcommand{\blue}[1]{\textcolor{blue}{\textbf{#1}}}

\usepackage[pagebackref,breaklinks,colorlinks]{hyperref}

\usepackage[capitalize]{cleveref}
\crefname{section}{Sec.}{Secs.}
\Crefname{section}{Section}{Sections}
\Crefname{table}{Table}{Tables}
\crefname{table}{Tab.}{Tabs.}

\def\cvprPaperID{*****} 
\def\confName{CVPR}
\def\confYear{2022}

\begin{document}

\title{CFA: Coupled-hypersphere-based Feature Adaptation \\for Target-Oriented Anomaly Localization}

\author{Sungwook Lee\\
Inha University\\
Incheon, South Korea\\
{\tt\small lsw2646@gmail.com}
\and
Seunghyun Lee\\
Inha University\\
Incheon, South Korea\\
{\tt\small lsh910703@gmail.com}
\and
Byung Cheol Song\\
Inha University\\
Incheon, South Korea\\
{\tt\small bcsong@inha.ac.kr}
}

\maketitle

\begin{abstract}
   For a long time, anomaly localization has been widely used in industries. Previous studies focused on approximating the distribution of normal features without adaptation to a target dataset. However, since anomaly localization should precisely discriminate normal and abnormal features, the absence of adaptation may make the normality of abnormal features overestimated. Thus, we propose Coupled-hypersphere-based Feature Adaptation~(CFA) which accomplishes sophisticated anomaly localization using features adapted to the target dataset. CFA consists of (1)~a learnable patch descriptor that learns and embeds target-oriented features and (2)~scalable memory bank independent of the size of the target dataset. And, CFA adopts transfer learning to increase the normal feature density so that abnormal features can be clearly distinguished by applying patch descriptor and memory bank to a pre-trained CNN. The proposed method outperforms the previous methods quantitatively and qualitatively. For example, it provides an AUROC score of 99.5\% in anomaly detection and 98.5\% in anomaly localization of MVTec AD benchmark. In addition, this paper points out the negative effects of biased features of pre-trained CNNs and emphasizes the importance of the adaptation to the target dataset. The code is publicly available at \url{https://github.com/sungwool/CFA_for_anomaly_localization}
\end{abstract}

\section{Introduction}
\label{sec:introduction}
    Anomaly detection is a well-known computer vision task to detect anomalous feature(s) in a given image. The human visual system~(HVS) can easily recognize unexpected patterns in images, i.e., anomalies, regardless of feature complexity. With the rapid development of CNNs, machine vision systems can recognize anomalies by learning abstract features. In addition to the image-level anomaly detection, anomaly localization, that is, pixel-level anomaly detection, has also been actively studied. Anomaly localization provides a heatmap indicating the location of an outlier as well as the presence or absence of the outlier. Note that the heatmap can be the starting point for explaining the cause of the anomaly.
    
    Meanwhile, anomaly localization algorithms cannot consider all possible outliers in learning. In other words, they cannot build a dataset that includes all outliers. So, distinguishing abnormal samples by learning the distribution of normal samples has been the mainstream approach. For example, unsupervised learning-based approaches such as \cite{SSIM-AD,anogan} utilized the characteristic that a generator trained with only normal features cannot successfully reconstruct abnormal features. Self-supervised learning-based approaches such as~\cite{cutpaste,anoseg,patchSVDD} synthesized noise and used them as abnormal samples in learning. Recently, ~\cite{SPADE,PaDiM,PatchCore} designed memory banks using pre-trained CNNs with large datasets such as ImageNet~\cite{imagenet} and achieved state-of-the-art~(SOTA) performance. This memory bank-based approach extracts sufficiently generalized features from a pre-trained CNN without learning the target dataset and then stores them into a memory bank. Finally, it determines whether an input sample is abnormal by matching the input features with the memorized features.
    
    However, industrial images generally have a different distribution from ImageNet. So, the pre-trained CNN extracts only unfitted features from new industrial images. This can be a fatal problem in anomaly localization, which requires a precise distinction between normal and abnormal features. \cite{PatchCore} pointed out the mismatch problem caused by pre-trained CNNs extracting biased features. It only used mid-level features with relatively small biases, but did not fundamentally solve the mismatch problem.
    
    The performance of anomaly localization depends on the size of the memory bank. Conventional methods stored as many normal features of the target dataset as possible in a memory bank to accommodate unfitted features, that is, to understand the distribution of normal features. So, the size of the memory bank was determined in proportion to that of the target dataset. However, a great number of unfitted features in the memory bank may cause the risk of overestimated normality of abnormal features. Furthermore, a large capacity memory bank increases the inference time.
    
    To obtain discriminative normal features, we propose a novel approach to produce target-oriented features with reduced bias by applying transfer learning to a pre-trained CNN. First, we define a novel loss function based on soft-boundary regression that searches a hypersphere with a minimum radius to densely cluster normal features. The proposed loss function helps the learnable patch descriptor extract discriminative features by utilizing several memorized features that form a coupled-hypersphere. Next, to reduce the inference time, we present a scalable memory bank. Since the scalable memory bank is independent of the size of the target dataset, it not only alleviates the risk of overestimated normality of abnormal features, but also achieves the efficiency of spatial complexity. Therefore, the proposed method can effectively localize anomalies by extracting appropriately target-oriented features to the target dataset and constructing a down-scaled memory bank to have core normal features.
    \begin{figure*}[t!] 
\begin{center}
    \includegraphics[width=0.95\linewidth]{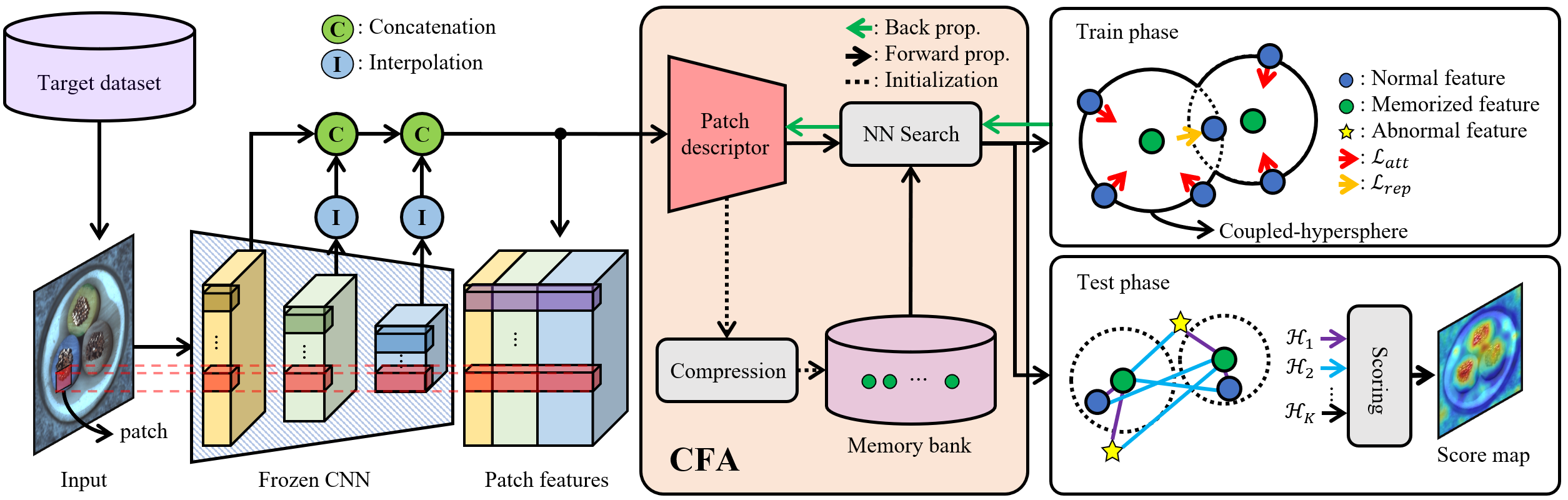}
\end{center}
\caption{Overall structure of our proposed method~(CFA).}\vspace{-0.15cm}
\label{fig:main}
\end{figure*}

 
    We evaluated the proposed method using MVTec AD benchmark~\cite{mvtec}, which is a popular industrial image dataset for visual inspection. The proposed method showed a performance of 99.5\% in terms of anomaly detection performance index, i.e., image-level AUROC (I-AUROC), and accomplished SOTA performance of 98.5\% in terms of anomaly localization performance index, i.e., pixel-level AUROC (P-AUROC). In particular, it is worth noting that the proposed method provides better performance than conventional methods while decreasing the activations of about 99.9\% of the memory bank~\cite{activations}.
    
    \textbf{Contributions.}~The contributions of this paper are summarized as follows: 1)~We discover the negative effects of biased features from pre-trained CNNs on anomaly localization, and propose an adaptation to the target dataset as a solution. 2)~We propose a new approach to acquire discriminative features through metric learning, and experimentally verify that the features enable very sophisticated anomaly localization. 3)~A memory bank that is compressed independently of the size of the target dataset through feature adaptation achieves SOTA performance despite its significantly reduced capacity.

\section{Related Works}
    In general, the acquisition of outlier samples requires a lot of costs and it is also impossible to consider all types of outliers. So, a memory bank-based approach that acquires normal features by inferring the target dataset using pre-trained CNNs has emerged. \cite{SPADE} obtained normal features from the feature maps and stored them in a memory bank. And during the test time, it calculated anomaly scores by computing the Euclidean distance between the normal features from the memory bank and the patch features from a test sample. \cite{PaDiM} defined a memory bank by modeling the normal distribution at each location of the feature map. To further consider the inter-feature correlation, it adopted Mahalanobis distance metric for computing anomaly scores. \cite{PatchCore} used only mid-level feature maps to mitigate biased features and maximized nominal information by considering the neighbor features of each normal feature. In addition, it proposed greedy coreset subsampling to lighten the memory bank and the time/space complexity. However, the above-mentioned methods have in common that they use features biased on a large dataset without adaptation. Also, the size of the memory bank is still proportional to that of the target dataset, and there is a problem that the memory bank cannot be adjusted to an arbitrary size.

\section{Proposed Method}
    This paper proposes a so-called Coupled-hypersphere-based Feature Adaptation~(CFA) that performs transfer learning on the target dataset as a solution to alleviate the bias of pre-trained CNNs. The patch descriptor of CFA learns the patch features obtained from normal samples of a target dataset to have a high density around the memorized features. Thus, CFA solves the problem that the normality of abnormal features is overestimated when using a pre-trained CNN.
    
    As in Fig.~\ref{fig:main}, CFA acquires feature maps of various scales by inferring samples of the target dataset based on a pre-trained CNN with a large dataset, that is, a biased CNN. Since the feature maps sampled at each depth of CNN have different spatial resolutions, they are interpolated to have the same resolution and then concatenated, as in \cite{PaDiM}. As a result, patch features~$\mathcal{F}\in\mathbb{R}^{D\times H \times W}$ are generated. Here, $H$ and $W$ mean the height and width of the largest features map, respectively, and $D$ indicates the sum of dimensions of the sampled feature maps. Since each pixel location of $\mathcal{F}$ has a predetermined receptive field, patch feature~$\mathbf{p}_{t\in \{1,\ldots,\textit{HW}\}}\in\mathbb{R}^{D}$ can be considered as semantic information at the pixel location.
    Next, $\mathbf{p}$ is input to the patch descriptor~$\phi(\cdot):\mathbb{R}^{D} \rightarrow \mathbb{R}^{D'}$. Here, $\phi(\cdot)$ is an auxiliary network with learnable parameters, which converts $\mathbf{p}_{t}$ into target-oriented features~$\phi(\mathbf{p}_ {t}) \in \mathbb{R}^{D'}$. Here, $D'$ means the dimension of $\phi(\mathbf{p}_{t})$ embedded by $\phi(\cdot)$.
    
    Meanwhile, all initial target-oriented features acquired from the train set consisting of only normal samples are stored in the memory bank~$\mathcal{C}$ according to a specific modeling procedure. In Fig.~\ref{fig:main}, the dotted line indicates that it is performed only in the initialization step~(c.f. section~\ref{section:memory-bank}). In the train phase, CFA performs contrastive supervision based on the superimposed hyperspheres created with the memorized features~$\mathbf{c} \in \mathcal{C}$ as the centers, that is, the so-called coupled-hypersphere. Note that $\phi(\mathbf{p}_{t})$s trained to be densely clustered in the train phase, i.e., normal features, are very useful for distinguishing abnormal features. In the test phase, CFA matches $\mathbf{p}_t$ obtained from the arbitrary sample of the test set with the nearest neighbor~$\mathbf{c}_t$ searched in the memory bank, and generates heatmaps representing the degree of anomality. Finally, a score map for anomaly localization from the heatmaps is calculated by a specific scoring function~(c.f. section~\ref{section:scoring-function}). Note that the score map shows the refined region of abnormal features.
    
    This paper is organized as follows: Section~\ref{section:patch-descriptor} defines $\phi(\cdot)$ and explains how to train it, and section~\ref{section:memory-bank} defines $\mathcal{C}$. Finally, section~\ref{section:scoring-function} shows the process of calculating anomaly scores.
    
    \subsection{Coupled-hypersphere-based Feature Adaptation} \label{section:patch-descriptor}
        This section describes how to learn $\phi(\cdot)$ attached to a pre-trained CNN through transfer learning based on a memory bank for more sophisticated target-oriented anomaly localization. Previous studies \cite{DeepSVDD} and \cite{patchSVDD} that learned the distribution of target datasets by introducing the hypersphere concept still had a problem in not clearly understanding normal features because they did not use a memory bank. Therefore, we present a method for effectively learning $\phi(\cdot)$. The proposed method can solve the bias problem of pre-trained CNNs by fusing a hypersphere-based loss function and a memory bank. The specific process is as follows:
        
        To obtain a feature space that can clearly detect abnormal features, we extract clustered normal features so that $\phi(\cdot)$ has a high density. First, the $k$-th nearest neighbor~$\mathbf{c}^{k}_{t}$ is searched through the NN search of $\phi(\mathbf{p}_t)$ and $\mathcal{C}$. Next, CFA supervises $\phi(\cdot)$ so that $\mathbf{p}_t$ is embedded close to $\mathbf{c}^{k}_{t}$.
        Specifically, $\phi(\cdot)$ makes it possible to form a high concentration between normal features by supervising $\mathbf{p}_t$ to embed it inside a hypersphere of radius~$r$ created with $\mathbf{c}^{k}_{t}$ as the center. So, $\mathcal{L}_{\textit{att}}$ for attracting by adding a penalty to $\phi(\mathbf{p}_t)$ as far as $r$ away from $\mathbf{c}^{k}_{t}$ is described by
        \begin{equation} \label{eq:positive-term}
            \mathcal{L}_{\textit{att}} = \frac{1}{T K} \sum^{T}_{t=1} \sum^{K}_{k=1} \max \{0, \mathcal{D}(\phi(\mathbf{p}_t), \mathbf{c}^k_t) - r^2\} 
        \end{equation}
        where a hyperparameter~$K$ is the number of nearest neighbors matching with $\phi(\mathbf{p}_t)$ and $T=h\times w$ is the number of $\mathbf{p}$s obtained from a single sample. $\mathcal{D}(\cdot, \cdot)$ is a predefined distance metric, i.e., Euclidean distance in this paper. Eq~(\ref{eq:positive-term}) induces $\phi(\mathbf{p}_t)$ to gradually approach the hypersphere created with $\mathbf{c}^{k}_{t}$ as the center. So, CFA enables feature adaptation by optimizing the parameters of $\phi(\cdot)$ to minimize $\mathcal{L}_{\textit{att}}$ through transfer learning. As such, if $\phi(\mathbf{p})$ is densely clustered through feature adaptation using $\mathcal{L}_{\textit{att}}$, it will be easy to distinguish from abnormal features.
        
        However, the ambiguous $\phi(\mathbf{p})$ belonging to multiple hyperspheres at the same time still leaves room for the normality of abnormal features to be overestimated. To address this, we additionally use hard negative features to perform contrastive supervision to obtain a more discriminative $\phi(\mathbf{p}_t)$. Hard negative features are defined as the $\textit{K+j}$-th nearest neighbor~$\mathbf{c}^j_t$ of $\mathbf{p}_t$ matched through NN search with $\mathcal{C}$. Thus, we define $\mathcal{L}_{\textit{rep}}$ that supervises $\phi(\cdot)$ contrastively so that the hypersphere created with $\mathbf{c}^j_t$ as the center repels $\mathbf{p}_t$ as follows.
        \begin{equation} \label{eq:negative-term}
            \mathcal{L}_{\textit{rep}} = \frac{1}{T J} \sum^{T}_{t=1} \sum^{J}_{j=1} \max \{0, r^2 - \mathcal{D}(\phi(\mathbf{p}_t), \mathbf{c}^j_t) - \alpha \} 
        \end{equation}
        where the hyperparameter~$J$ is the total number of hard negative features to be used for contrastive supervision and the hyperparameter~$\alpha$ is used to control the balance between $\mathcal{L}_{\textit{att}}$ and $\mathcal{L}_{\textit{rep}}$.

        As a result, CFA optimizes the parameters of $\phi(\cdot)$ through transfer learning using Eqs.~(\ref{eq:positive-term}) and (\ref{eq:negative-term}) together:
        \begin{equation} \label{eq:total-term}
            \mathcal{L}_{\textit{CFA}} = \mathcal{L}_{\textit{att}} + \mathcal{L}_{\textit{rep}}
        \end{equation}
        If the distance between $c^j_t$ and $c^k_t$ matched with $\mathbf{p}_t$ is closer than $r$, $\mathcal{L}_{\textit{CFA}}$ of Eq.~{\ref{eq:total-term}} directly supervises $\mathbf{p}_t$ based on the coupled-hypersphere. Thus, $\mathbf{p}_t$ can be embedded by $\phi(\cdot)$ so that the hypersphere of $\mathbf{c}^k_t$ has a higher density through contrastive supervision using $\mathbf{c}^j_t$. Note that we named the process of obtaining target-oriented features from the patch descriptor through transfer learning using $\mathcal{L}_{\textit{CFA}}$ 'Coupled-hypersphere-based Feature Adaptation'.
        
    \subsection{Memory Bank Compression} \label{section:memory-bank}
        Transfer learning through the proposed CFA requires a memory bank for effective adaptation to target dataset. However, as seen in Table~\ref{table:complexity}, the complexity of the modeling process or memory bank space in the previous methods~\cite{SPADE,PaDiM,PatchCore} tends to increase in proportion to the size of the target dataset, i.e., $|\mathcal{X}|$. To mitigate this phenomenon, this section presents a compression scheme to construct an efficient memory bank.
        \begin{figure}[t!]
\begin{center}
    \includegraphics[width=1.0\linewidth]{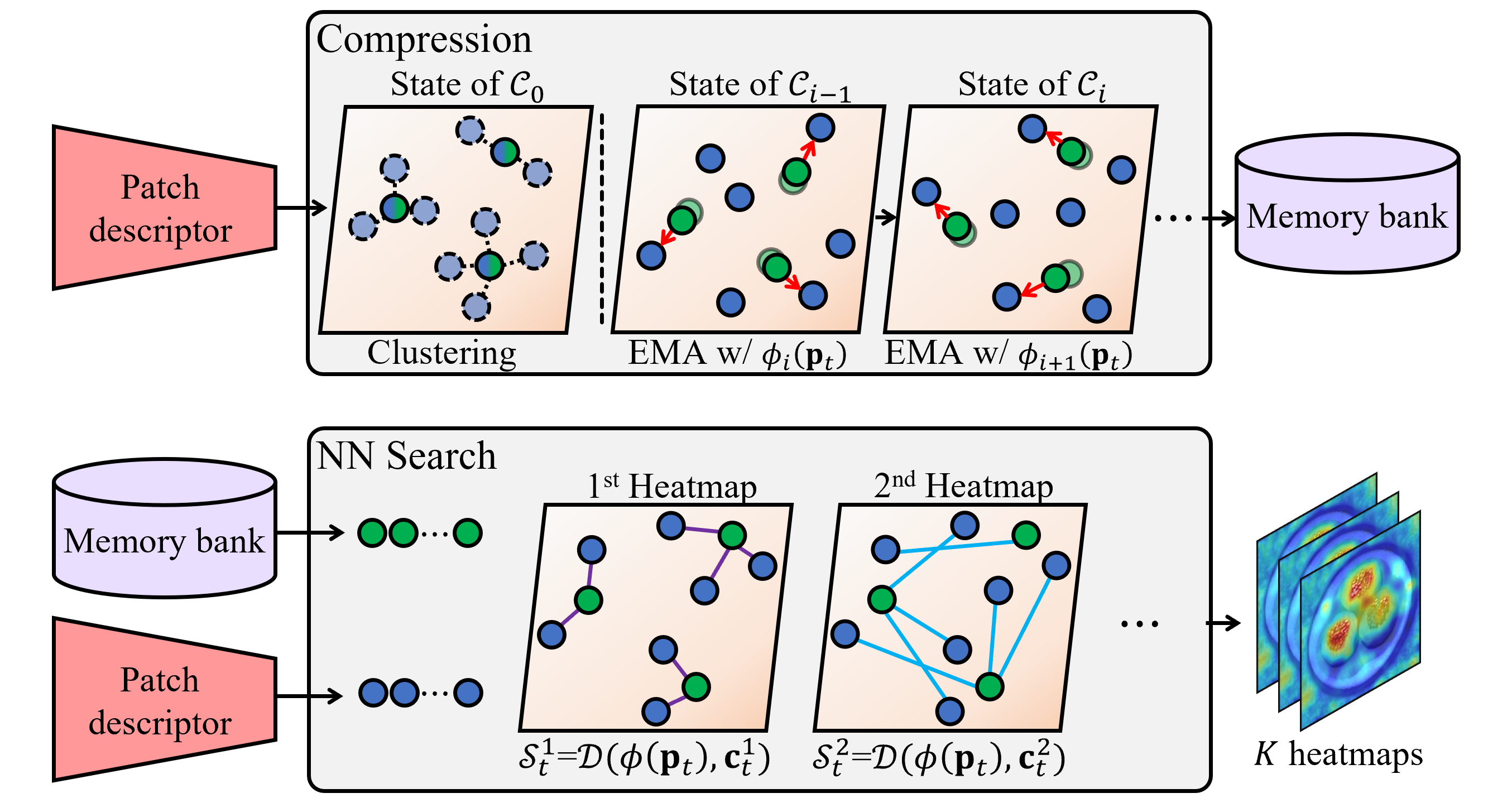}
\end{center}
\caption{(Upper) The process of initially modeling the memory bank (lower) the process of generating heatmaps through feature matching.}\vspace{-0.15cm}
\label{fig:method}
\end{figure}


        The compression process of the memory bank is described by algorithm~\ref{algorithm:modeling}. First, an initial memory bank~$\mathcal{C}_0$ is constructed by applying K-means clustering to all $\phi_{0}(\mathbf{p}_{t\in\{1,\ldots,T\}})$ obtained from the first normal samples~$\mathbf{x}_0$ of the train set~$\mathcal{X}$. The process of updating the memory bank after $\mathcal{C}_0$ is as follows: Infer the $i$-th normal sample~$\mathbf{x}_i$ and search for the set of nearest patch features~$\mathcal{C}^{\textit{NN}}_{i}$ from the $\textit{i-}1$-th memory bank~$\mathcal{C}_{\textit{i-}1}$. Next, the $i$-th memory bank of the next state~$\mathcal{C}_{i}$ is calculated by exponential moving average~(EMA) of $\mathcal{C}^{\textit{NN}}_{i}$ and $\mathcal{C}_{\textit{i-}1}$. The final memory bank $\mathcal{C}$ is obtained by repeating the above process $|\mathcal{X}|$ times for all normal samples of the train set.
        
        The upper part of Fig.~\ref{fig:method} illustrates the process in which memorized features are updated for each sample of the target dataset. Unfortunately, $\mathcal{C}_0$ initialized through the $K$-NN search does not represent $\mathcal{X}$ as a whole. However, if $\mathcal{C}_{\textit{i-1}}$ is updated through EMA iteratively along $\phi_{i}(\mathbf{p}_t)$, the final $\mathcal{C}$ can store the core normal features representing $\mathcal{X}$.
        
        Since the proposed algorithm~\ref{algorithm:modeling} updates $\mathcal{C}$ in every state, the modeling process requires the space complexity as much as $\mathcal{O}(HWD')$. Also, $\mathcal{C}$ has $\phi(\mathbf{p})$ of feature dimension $D'$ as many as the number of cluster centers, so it has as much spatial complexity as $\mathcal{O}(\gamma(HW)D')$. Here, the compression ratio~$\gamma$ indicates the ratio of $T$, i.e., the number of $\mathbf{p}$s obtained from $\mathcal{F}$ and the number of cluster centers. Therefore, $\mathcal{C}$ of the proposed method is not affected by $|\mathcal{X}|$ as in Table~\ref{table:complexity}.
        
    \subsection{Scoring Function} \label{section:scoring-function}
        In Section~\ref{section:patch-descriptor}, the distance between $\phi(\mathbf{p}_t)$ and $\mathbf{c}^k_t$ was calculated using $\mathcal{D}(\cdot, \cdot)$. $\min_k\mathcal{D}(\phi(\mathbf{p}_t), \mathbf{c}^k_t)$ means the minimum distance between $\phi(\mathbf{p}_t)$ and the memorized features $\mathcal{C}$, that is, the degree of anomality of $\phi(\mathbf{p}_t)$. So, we can define the anomaly score naively using $\mathcal{D}(\phi(\mathbf{p}_t), \mathbf{c}^k_t)$ as it is, as shown below:
        \begin{equation} \label{eq:simple-score}
            \mathcal{S}_{t} =\min_k \mathcal{D}(\phi(\mathbf{p}_t), \mathbf{c}^{k}_{t})
        \end{equation}
        However, since normal features are continuously distributed, the boundaries between clusters are not clear. So, it is difficult to discriminate abnormal features with the naive anomaly score precisely. In detail, it can be uncertain which memorized features will match $\phi(\mathbf{p}_t)$.
        In this case, even though $\phi(\mathbf{p}_t)$ is a normal feature, it exists in the middle of the memorized features, resulting in a large distance. So, the naive anomaly score based only on distance risks underestimates the normality of normal features. Thus, we propose a novel scoring function that considers the certainty of $\phi(\mathbf{p}_t)$.

\begin{algorithm}[t]
\caption{Memory Bank Modeling.}
    \begin{algorithmic}\small
        \STATE \textbf{Require\textbf{}}:Patch descriptor $\phi$, dataset $\mathcal{X}$, EMA parameter $\beta$
        \STATE \textbf{Initialization}: $\mathcal{C}_{0} \gets \text{KMeans}{\phi_{0}(\textbf{p}})$
        \FOR{ $i \in \{1, \ldots, |\mathcal{X}|\}$ }
            \STATE $\mathcal{C}^{\textit{NN}}_{i} \gets \{\}$
            \FOR{ $j \in \{1, \ldots, |\mathcal{C}|\}$ }
                \STATE $Y \gets (\phi_{i}(\textbf{p}) \cup \mathcal{C}^{\textit{NN}}_{i}) \cap (\mathcal{C}^{\textit{NN}}_{i})^{c}$
                \STATE $\mathcal{C}^{\textit{NN}}_{i} \cup \arg \min_{y \in Y}{\norm{y - \mathcal{C}^{j}_{i-1}}_{2}}$
            \ENDFOR
            \STATE $\mathcal{C}_{i} \gets \ (1-\beta) \cdot \mathcal{C}_{i-1} + \beta \cdot \mathcal{C}^{\textit{NN}}_{i} $
        \ENDFOR
        \STATE $\mathcal{C} \gets \mathcal{C}_{|\mathcal{X}|}$
        \RETURN $\mathcal{C}$
    \end{algorithmic}
\label{algorithm:modeling}
\end{algorithm}
        \begin{table}[t!] 
\caption{Complexity estimates of memory bank modeling and memory bank size.}
\begin{center}\small
\begin{tabular}{c | c | c}
    \toprule
    \textbf{Methods} & Modeling & Memory Bank \\
    \midrule
    SPADE      & $\mathcal{O}(|\mathcal{X}|HWD) $ & $\mathcal{G} \in \mathbb{R}^{|\mathcal{X}| \times H\times W\times D}$\\
    \midrule
    PaDiM      & $\mathcal{O}(|\mathcal{X}|HWD^2) $ & $\mathcal{N}(\mu, \boldsymbol{\Sigma}) \in \mathbb{R}^{H\times W\times D^2}$\\  
    \midrule
    PatchCore  & $\mathcal{O}(|\mathcal{X}|HWD') $ & $\mathcal{M} \in \mathbb{R}^{|\mathcal{X}| \times \gamma (H\times W)\times D'}$\\  
    \midrule
    Ours       & $\mathcal{O}(HWD') $ & $\mathcal{C} \in \mathbb{R}^{\gamma (H\times W\times D)}$\\
    \bottomrule
\end{tabular}
\end{center}
\label{table:complexity}\vspace{-0.15cm}
\end{table}
        \begin{table*}[t!]
\caption{Image/Pixel-level AUROC (\%) of anomaly localization methods on MVTec AD dataset.}
\begin{center}
\begin{tabular}{c | c | c c c c c c | c c}
\toprule
\multicolumn{2}{c|}{Model}           & SPADE & Patch SVDD & PaDiM & CutPaste & CFLOW & PatchCore & CFA & CFA++ \\
\midrule
\multirow{3}{*}{I-AUROC}  & Textures &  96.6 &    94.5    &  95.3 &   98.4   &  98.7 &    99.0   & 99.6 & \textbf{99.8}  \\
                          & Objects  &  96.0 &    90.8    &  95.3 &   94.1   &  98.0 &    99.1   & 99.2 &  \textbf{99.4}  \\
                          &   All    &  96.2 &    92.1    &  95.3 &   95.5   &  98.3 &    99.1   & 99.3 &  \textbf{99.5}  \\
\midrule
\multirow{3}{*}{P-AUROC}  & Textures &  92.9 &    93.7    &  95.3 &   96.9   &  \textbf{98.5} &    97.5   & 97.2 &  97.5  \\
                          & Objects  &  97.6 &    96.7    &  95.3 &   97.8   &  98.7 &    98.3   & 98.6 &  \textbf{98.9}  \\
                          &   All    &  96.0 &    95.7    &  97.5 &   97.5   &  \textbf{98.6} &    98.2   & 98.2 &  98.5  \\
\bottomrule
\end{tabular}
\end{center}
\label{table:mvtec-ad}\vspace{-0.15cm}
\end{table*}

        The clearer $\phi(\mathbf{p}_t)$ is matched, the closer the distance to a specific memorized feature is compared to other memorized features. Thus, we use softmin to measure how close the nearest $c$ is compared to the other $c$, and define it as certainty. As a result, the problem of underestimated normality is figured out by multiplying $\mathcal{S}^k_t$ with the certainty of $\phi(\mathbf{p}_t)$. The formulation is described as follows:
        \begin{equation} \label{eq:score}
            \mathcal{A}_{t} = \frac{e^{-\mathcal{S}_{t}}}{\sum^{K}_{k=1}e^{-\mathcal{D}(\phi(\mathbf{p}_t), \mathbf{c}^{k}_{t})}} \cdot {\mathcal{S}_{t}}
        \end{equation}
        Finally, in the test phase of CFA, the anomaly score map, which is the final output of anomaly localization, is obtained from the heatmaps. Note that the heatmaps are generated from naive anomaly scores, which is illustrated in the lower part of Fig.~\ref{fig:method}. Briefly, the $k$-th heatmap~$\boldsymbol{\mathcal{H}}^k = \{\mathcal{D}(\phi(\mathbf{p}_t), \mathbf{c}^k_t) | 1\leq t \leq T\}$ is generated and rearranged so that $\boldsymbol{\mathcal{H}}^k$ has spatial information. Then, Eq.~(\ref{eq:score}) is calculated at all pixel locations to obtain the final output of CFA, i.e., anomaly score map~$\boldsymbol{\mathcal{A}}$. Here, in order to output the anomaly score map with the same resolution as the input samples, $\boldsymbol{\mathcal{A}}$ is properly interpolated, and Gaussian smoothing of $\sigma = 4$ is applied as post-processing.
        
        In summary, CFA performs transfer learning for target-oriented anomaly localization using the proposed patch descriptor and memory bank. Then, CFA generates heatmaps from task-oriented features and computes sophisticated anomaly scores from them. Therefore, CFA solves the problem that the normality of abnormal features caused by the biased features of the pre-trained CNN is overestimated.
    
\section{Experiments}
    This section presents various experimental results to evaluate the anomaly detection and localization performance of CFA.
    All the experiments were performed on MVTec AD benchmark~\cite{mvtec}, that is, the most famous dataset in the anomaly localization field. 
    To verify the robustness of the proposed method, we also presented the performance for Rd-MVTec AD dataset which randomly rotated and cropped MVTec AD dataset.
    As an evaluation metric, we adopted Area Under the Receiver Operator Curve~(AUROC), and then evaluated the performance of the proposed method in terms of anomaly detection~(I-AUROC) and localization~(P-AUROC). In some experiments, we used Area Under the Per-Region-Overlap curve~(P-AUPRO)~\cite{aupro} which can evaluate anomaly localization more precisely.
    
    \subsection{Experimental setup}
        This section describes the configurations set up for experiments in this paper. 
        All CNNs used in the experiments were pretrained with ImageNet.
        To secure multi-scale features on pre-trained CNN, we extract feature maps corresponding to $\{C_2, C_3, C_4\}$ from intermediate layers as in \cite{fpn}.
        The spatial resolution of each extracted feature map is 1/4, 1/8, and 1/16 of the input sample.
        Exceptionally, for EfficientNet, which has a very small channel dimension, several feature maps were used for each scale, which were divided into channel dimension values.
        A $1\times1$ CoordConv. layer~\cite{coordconv} was used as a Patch descriptor, and its parameters are initialized to He's initializer~\cite{he-init}. 
        To optimize parameters of patch descriptor, the AdamW~\cite{adamw} was used, and amsgrad~\cite{amsgrad} was applied. 
        Here, the learning rate was set to 1e-3 without any scheduler and weight decay was set to 5e-4. 
        The batch size was set to 4.
        Patch descriptor was trained for 30 epochs which take about 10 minutes per sub-class.
        As hyperparameters of CFA, $r$ and $\alpha$ were set to 1e-5 and 1e-1, respectively.
        As the number of nearest neighbors for each patch feature, $K$ and $J$ were equally set to 3.
        The GPU was Quadro RTX 5000, and the CPU was Intel(R) Xeon(R) CPU E5-2650 v4 @ 2.20 GHz to measure the throughput of the proposed method.
        
        The MVTec AD dataset used in the experiment is the largest dataset consisting of 5354 industrial samples, of which 1725 are test samples.
        It is divided into 15 sub-classes, and we perform transfer learning independently for each class.
        For pre-processing, each sample of the dataset is resized into $256\times256$, and is center-cropped into $224\times224$. 
        And we use the RD-MVTec AD dataset to further consider unaligned samples, which are more difficult to detect outliers.
        Each sample of the RD-MVTec AD dataset is rotated randomly within $\pm10^\circ$. 
        After a random rotation, each sample is resized to $256\times 256$ and then randomly cropped to $224\times224$. 
        \begin{table}[t]
\caption{Image-level AUROC~(\%) and Pixel-level AUPRO~(\%) of anomaly localization methods on RD-MVTec AD dataset.}
\begin{center}\small
\begin{tabular}{c | c | c c c}
    \toprule
    \multicolumn{2}{c|}{Model}             & Textures & Objects & All  \\
    \midrule
    VAE                         & I-AUROC  &   54.7   &   65.8  & 62.1 \\
    (ResNet18)                  & P-AUPRO  &   23.1   &   30.2  & 27.8 \\    
    \midrule
    CFA++                       & I-AUROC  &   \textbf{98.6}   &   \textbf{95.5}  & \textbf{96.5} \\
    (ResNet18)                  & P-AUPRO  &   \textbf{81.1}   &   \textbf{82.2}  & \textbf{81.8} \\
    \midrule
    \midrule
    SPADE                       & I-AUROC  &   84.6   &   88.2  & 87.2 \\
    (WRN50-2)                   & P-AUPRO  &   75.6   &   65.8  & 69.0 \\    
    \midrule
    PaDiM                       & I-AUROC  &   92.4   &   92.1  & 92.1 \\
    (WRN50-2)                   & P-AUPRO  &   77.8   &   70.8  & 73.1 \\    
    \midrule
    CFA++                      & I-AUROC  &   \textbf{99.7}   &   \textbf{98.3}  & \textbf{98.7} \\
    (WRN50-2)                   & P-AUPRO  &   \textbf{82.2}    &  \textbf{83.7}  & \textbf{83.2} \\   
    \bottomrule
\end{tabular}
\end{center}
\label{table:rd-mvtec-ad}\vspace{-0.15cm}
\end{table}

        \begin{table*}[t!]
\caption{Performance comparison of image-level AUROC (\%) on each class of MVTec AD dataset. \red{Red}, \blue{blue}, and \textbf{bold} stand for the first, second, and third places  }
\begin{center}\small
\begin{tabular}{c | c c c c c c | c c}
    \toprule
    Class     & SPADE & Patch SVDD & PaDiM & CutPaste & CFLOW & PatchCore & CFA & CFA++\\
    \midrule
    Bottle    & -     &   98.6 &  -  & 98.2 &  \red{100}   & \red{100.0} & \red{100.0} & \red{100.0} \\
    \midrule
    Cable     & -     &   90.3 &  -  & 81.2 & 97.6 &\textbf{99.5} &  \red{99.8} &  \red{99.8} \\
    \midrule
    Capsule   & -     &   76.7 &  -  & \textbf{98.2} & 97.7 & \textbf{98.1} &  97.3 &  \red{99.2} \\
    \midrule
    Carpet    & -     &   92.9 &  -  & 93.9 & \blue{98.7} &\blue{98.7} &  97.3 &  \red{99.5} \\
    \midrule
    Grid      & -     &   94.6 &  -  & \red{100.0} & \textbf{99.6} & 98.2 &  99.2 &  \blue{99.9} \\
    \midrule
    Hazelnut  & -     &   92.0 &  -  & 98.3 & \red{100.0} & \red{100.0} &  \red{100.0} &  \red{100.0} \\
    \midrule
    Leather   & -     &   90.9 &  -  & \red{100.0} & \red{100.0} &\red{100.0} &  \red{100.0} &  \red{100.0} \\
    \midrule
    Metal nut & -     &   94.0 &  -  & 99.9 & 99.3 & \red{100.0} & \red{100.0} & \red{100.0} \\
    \midrule
    Pill      & -     &   86.1 &  -  & 94.9 & \textbf{96.8} & 96.6 &  \red{97.9} &  \red{97.9} \\
    \midrule
    Screw     & -     &   81.3 &  -  & 88.7 & 91.9 & \red{98.1} &  \blue{97.3} &  \blue{97.3} \\
    \midrule
    Tile      & -     &   97.8 &  -  & 94.6 & \blue{99.9} & 98.7 &  \textbf{99.4} & \red{100.0} \\
    \midrule
    Toothbrush& -     &  \red{100.0} &  -  & 99.4 & 99.7 & \red{100.0} & \red{100.0} & \red{100.0} \\
    \midrule
    Transistor& -     &   91.5 &  -  & 96.1 & 95.2 & \red{100.0} & \red{100.0} & \red{100.0} \\
    \midrule
    Wood      & -     &   96.5 &  -  & 99.1 & 99.1 & \textbf{99.2} &  \red{99.7} &  \red{99.7} \\
    \midrule
    Zipper    & -     &   97.9 &  -  & \red{99.9} & 98.5 &  99.4 &  \blue{99.6} &  \blue{99.6} \\
    \midrule
    \midrule
    Average   &   96.2 &  92.1 &  95.3 &  95.5 &  98.3 &  \textbf{99.1} &  \blue{99.3} & \red{99.5}\\
    \bottomrule
\end{tabular}
\end{center}
\label{table:details}\vspace{-0.15cm}
\end{table*}

    \subsection{Quantitative Results}
        This section investigates the quantitative performance of CFA. Table~\ref{table:mvtec-ad} shows I-AUROC and P-AUROC of CFA on MVTec AD dataset when used WRN50-2 pretrained with ImageNet. Here, CFA++ refers to a case that ensembles the results when using cropped images and using only resized samples. The proposed method provided SOTA performance for both texture classes and object classes in terms of I-AUROC, which evaluates the image-level anomaly detection performance. For instance, CFA++ shows 0.4~\% better I-AUROC than PatchCore~\cite{PatchCore} which has shown SOTA performance so far. Also, even in the aspect of P-AUROC, which is an evaluation metric for pixel-level anomaly localization, the proposed method achieved SOTA performance for object classes. But, the proposed method has slightly lower performance than CFLOW~\cite{cflow} in terms of P-AUROC when all classes are considered. Nonetheless, while the conventional methods show strength only in a one of anomaly detection and localization, the proposed method guarantees excellent performance in both scenarios. Also, note that the proposed method achieves outstanding performance through feature adaptation despite using a memory bank with a smaller spatial complexity compared to SPADE, PaDiM and PatchCore. We show the performance of each architecture in Table~\ref{table:details} to demonstrate the performance improvement effect of the proposed method more clearly. We can find that CFA++ provides the highest performance in most classes. In particular, the worst performance of both CFA and CFA++ is just 97.3\%, which is much higher than most other techniques. This tendency is because CFA has generalized performance to various classes due to the effect of the proposed feature adaptation.
        

\begin{table}[t!]
\caption{Image/Pixel-level AUROC (\%) of the proposed method according to $\mathcal{L}_{att}$ and $\mathcal{L}_{rep}$ on MVTec AD dataset.}
\begin{center}\small
\begin{tabular}{c | c c | c c}
    \toprule
    Backbone                   &$\mathcal{L}_{att}$&$\mathcal{L}_{rep}$& I-AUROC & P-AUROC  \\
    \midrule
    \multirow{3}{*}{ResNet18}  &            &            &  83.7   &   92.4   \\    
                               &$\checkmark$&            &  97.8   &   97.8   \\  
                               &$\checkmark$&$\checkmark$&  \textbf{98.9}   &   \textbf{98.1}   \\  
    \midrule
    \multirow{3}{*}{WRN50-2}   &            &            &  85.9   &   94.0   \\    
                               &$\checkmark$&            &  99.1   &   98.3   \\   
                               &$\checkmark$&$\checkmark$&  \textbf{99.5}   &   \textbf{98.5}   \\   
    \bottomrule
\end{tabular}
\end{center}
\label{table:learning-ablation}\vspace{-0.15cm}
\end{table}

        Table~\ref{table:rd-mvtec-ad} shows the performance of CFA on RD-MVTec AD dataset.
        The RD-MVTec AD dataset consists of the same samples as the MVTec AD dataset, but is not aligned.
        So, the performance for this dataset is generally lower than that for MVTec AD. Comparing with Table~\ref{table:rd-mvtec-ad}, for example, SPADE's I-AUROC fell by 9\%. This means that SPADE is very vulnerable to wild dataset.
        On the other hand, even in unaligned samples, I-AUROC of CFA++ showed marginal performance degradation of 0.8\%, which indicates that the proposed method can distinguish normal features as robustly as HVS.
        Also, compared to SPADE and PaDiM, CFA++ showed 11.5\% and 6.6\% higher I-AUROC scores, respectively. A similar trend is also observed in terms of P-AUPRO for evaluating sophisticated detection. For example, CFA++ showed 10.1\% higher P-AUPRO score than PaDiM.
        
    \subsection{Ablation Study}
        Table~\ref{table:learning-ablation} shows the effect of feature adaptation on anomaly localization. First, take a look at the case of using only the biased features of the pre-trained CNN. 
        Non-adapted pre-trained CNN showed low performance due to the biased features even though they have rich features obtained from large dataset.
        This is because the normality of normal features was underestimated due to biased features, which negatively affected the anomaly localization. At this time, when only $\mathcal{L}_{\textit{att}}$ was used, CFA improved I-AUROC and P-AUROC scores up to 14.1\% and 5.4\% in the case of ResNet18. For WRN50-2, I-AUROC and P-AUROC scores were increased by 13.2\% and 4.3\%, respectively, thanks to $\mathcal{L}_{\textit{att}}$. This is because normal features are more densely clustered around memorized features. However, a problem remains that they are not discriminative as it is still uncertain which hypersphere they belong to. Therefore, using $\mathcal{L}_{\textit{rep}}$ introduced to obtain discriminative features, further performance improvement is expected. In fact, in ResNet18, I-AUROC and P-AUROC scores improved by 1.1\% and 0.3\%, respectively, and in WRN50-2, they improved by 0.4\% and 0.2\%, respectively. By inducing normal features to be clustered more discriminatively, abnormal features were more precisely differentiated. On the other hand, it is interesting that ResNet18 shows a greater performance improvement than WRN50-2. Since ResNet18 uses a relatively small feature dimensions which may increase ambiguity, the problem of hypersphere overlapping can be severed. $\mathcal{L}_{\textit{rep}}$ effectively solved this problem.
        
        \begin{table}[t!]
\caption{Pixel-level AUROC (\%) of the proposed method with additional memory bank compression on MVTec AD dataset.}
\begin{center}\small
\begin{tabular}{c | c c | c | c}
    \toprule
    Backbone                   & $\gamma_d$ & $\gamma_c$ & P-AUROC  & Throughput \\
    \midrule
    \multirow{4}{*}{WRN50-2}   &    1       &    1       &   \textbf{98.45}  &  48  \\  
                               &    1/2     &    1/2     &   98.44  &  93~(1.9x)    \\    
                               &    1/4     &    1/4     &   98.44  &  132~(2.8x)   \\  
                               &    1/8     &    1/8     &   98.36  &  \textbf{172}~(3.6x)\\  
    \bottomrule
\end{tabular}
\end{center}
\label{table:compression-ablation}\vspace{-0.15cm}
\end{table}
        Table~\ref{table:compression-ablation} shows P-AUROC scores when the memory bank is compressed by additionally employing feature dimension reduction ratio~$\gamma_d$ and patch reduction ratio~$\gamma_c$. First, note that the memory bank of CFA has a size independent of the target dataset. For example, in the Bottle class consisting of 209 samples, CFA was compressed to a size of approximately $\frac{1}{\mathcal{|X_\textit{bottle}|}}$ or 0.5\%. That is, the compression ratio~$\gamma$ of each sub-class is calculated as $\frac{\gamma_d \gamma_c}{|\mathcal{X}|}$. Even in the memory bank compressed from 25\% to about 2\%, the P-AUROC score of CFA was slightly decreased by 0.08\%.
        
        The use of such a lightweight memory bank has a positive effect on the increase in throughput. The throughput of CFA considers inference times of forward pass through pre-trained CNN and patch descriptor. We can observe that if the memory bank is further compressed in the same experimental environment, the throughput increases up to 3.6 times. For example, looking at the 3rd row of Table~\ref{table:compression-ablation}, even if the activation of the memory bank is reduced by about 99.9\%, the CFA performance hardly decreases and the throughput rather increases up to 2.8 times. This is because the memory bank is compressed to extract only the core features of $\mathcal{X}$, and adaptation is performed so that these features are densely clustered.
        
        \begin{table}[t!]
\caption{Image/Pixel-level AUROC (\%) of the proposed method with various pretrained CNNs on MVTec AD dataset.}
\begin{center}\small
\begin{tabular}{c | c | c c}
    \toprule
    Backbone                           &      Method     & I-AUROC & P-AUROC  \\
    \midrule
    \multirow{2}{*}{VGG19}             &       DFR       &  93.8   &   \textbf{95.5}   \\    
                                       &     CFA++       &  \textbf{96.2}   &  95.3    \\ 
    \midrule
    \multirow{2}{*}{EffiNet-B5}        &    PaDiM        &  97.9   &     97.5    \\
                                       &     CFA++       &  \textbf{98.8}  &    \textbf{98.0}    \\  
    \midrule
    \multirow{2}{*}{ResNet18}          &    CFLOW      &  96.8   &   \textbf{98.1}   \\    
                                       &     CFA++      &  \textbf{98.9}   &   \textbf{98.1}   \\ 
    \bottomrule
\end{tabular}
\end{center}
\label{table:backbone-ablation}\vspace{-0.15cm}
\end{table}
        Table~\ref{table:backbone-ablation} shows the performance of anomaly detection and localization according to pre-trained CNNs. Here, CFA and previous methods~\cite{differnet,PaDiM,cflow} was compared for VGG19~\cite{vgg}, EffiNet-B5~\cite{efficientnet} and ResNet18~\cite{resnet}, which are most popularly used for anomaly localization. CFA showed superior I-AUROC scores by 2.4\%, 0.9\%, and 2.1\% than three pre-trained CNNs, respectively. Thus, Table~\ref{table:backbone-ablation} supports the universality of the proposed method.
        
    \subsection{Qualitative Results}
        \begin{figure}[t!]
\begin{center}
    \includegraphics[width=1\linewidth]{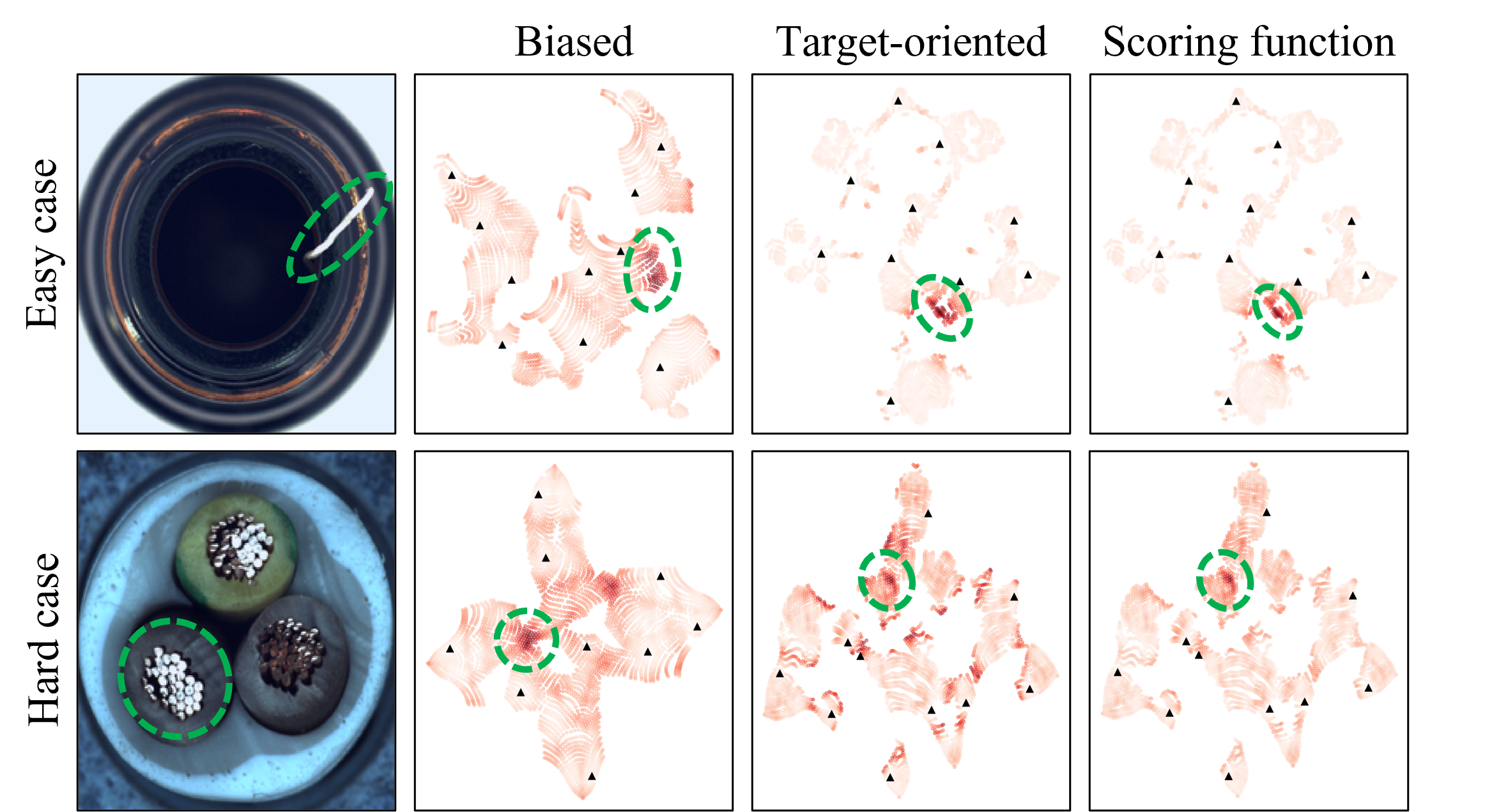}
\end{center}
\caption{Visualization of anomaly score of each patch feature.}
\label{fig:exp}\vspace{-0.15cm}
\end{figure}
        Fig.~\ref{fig:exp} shows the anomaly score of the patch features per sample according to feature adaptation and scoring function. Here, redness means anomaly score, a dotted circle means the area of abnormal features, and a triangle means memorized features. When a feature biased to a large dataset is used before adaptation, the normality of the normal feature is underestimated and has a score similar to that of the abnormal feature~(see the second column of Fig.~\ref{fig:exp}). It is difficult to distinguish the two features because the boundary is ambiguous in terms of score. This induces a negative effect that abnormal features cannot be precisely distinguished.
        \begin{figure*}[t!]
    \begin{center}
        \includegraphics[width=0.8\linewidth]{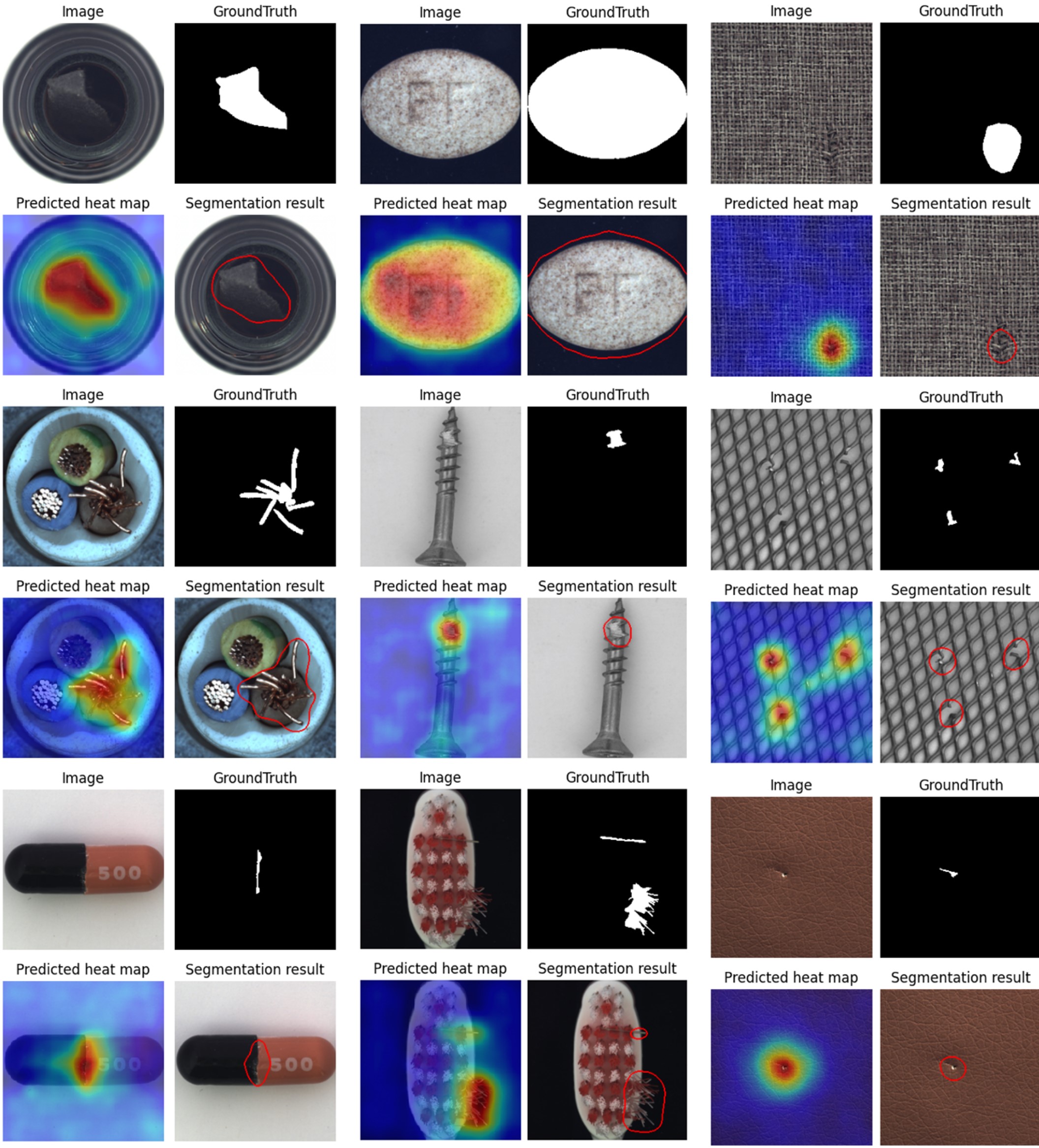}
    \end{center}
    \caption{Visualization of results of anomaly localization for object classes in MVTec AD benchmark.}
    \label{fig:object}\vspace{-0.15cm}
\end{figure*}
        
        On the other hand, when target-oriented features after feature adaptation are used, they are well clustered. So the normal features of the easy case and abnormal features are clearly distinguished~(see the third column of Fig.~\ref{fig:exp}). Still, clustering alone cannot precisely score the uncertain abnormal features of the hard case. The proposed scoring function determines the anomaly score by considering the certainty, so even the abnormal features of the hard case can be distinguished precisely, as shown in the last column of Fig.~\ref{fig:exp}. As a result, each step of the proposed method effectively improves the anomaly localization performance.
        
        Next, Fig.~\ref{fig:object} shows results of anomaly localization that indicate the abnormal areas.
        The anomaly score map obtained through CFA is interpolated to have the spatial resolution of the input sample and Gaussian filtered with $\sigma=4$ for smooth boundaries.
        Also, min-max scaling is performed for the normalized anomaly score.
        The threshold for segmentation result is obtained by calculating the F1-score for all anomaly scores of each sub-class. Experimental results prove that the proposed method can localize abnormal areas well even in rather difficult cases. In addition, we can find that the proposed method has consistent performance in both object and texture classes. As a result, the proposed method performs qualitatively as well.

\section{Conclusion}
    In this paper, we pointed out the bias problem caused by pre-trained CNNs in anomaly localization that mainly uses industrial images. To solve this problem, we proposed Coupled-Hypersphere-based Feature Adaptation~(CFA) to obtain target-oriented features. CFA consists of a learnable patch descriptor used with a pre-trained CNN and a memory bank storing memorized features. Through transfer learning and the feature adaptation of patch descriptor associating with a predetermined memory bank, CFA achieved successful target-oriented anomaly localization. CFA showed SOTA performance on the MVTec AD benchmark, the most representative dataset composed of industrial images. Then, the effectiveness of feature adaptation to the target dataset was examined qualitatively/quantitatively through extensive experiments.

{\small
\bibliographystyle{ieee_fullname}
\bibliography{egbib}
}

\end{document}